\documentclass[10pt,twocolumn,letterpaper]{article}

\usepackage{cvpr}              

\usepackage{graphicx}
\usepackage{amsmath}
\usepackage{amssymb}
\usepackage{booktabs}
\usepackage{amsfonts}
\usepackage{ulem}
\usepackage{multirow}
\usepackage[table,xcdraw]{xcolor}
\usepackage[accsupp]{axessibility}
\usepackage{pifont}
\newcommand{\cmark}{\ding{51}}%
\newcommand{\xmark}{\ding{55}}%

\usepackage[pagebackref,breaklinks,colorlinks]{hyperref}

\usepackage[capitalize]{cleveref}
\crefname{section}{Sec.}{Secs.}
\Crefname{section}{Section}{Sections}
\Crefname{table}{Table}{Tables}
\crefname{table}{Tab.}{Tabs.}

\def\cvprPaperID{3024} 
\def\confName{CVPR}
\def\confYear{2023}

\begin{document}

    \title{CORA: Adapting CLIP for Open-Vocabulary Detection with 

Region Prompting and Anchor Pre-Matching}


\author{
    Xiaoshi Wu$^{1}$, 
    Feng Zhu$^{2}$, 
    Rui Zhao$^{2, 3}$, 
    Hongsheng Li$^{1, 4}$
\vspace{0.1em}\\
    $^{1}$Multimedia Laboratory, The Chinese University of Hong Kong \\
    $^{2}$SenseTime Research \quad $^{3}$Qing Yuan Research Institute, Shanghai Jiao Tong University \\
    $^{4}$Centre for Perceptual and Interactive Intelligence (CPII) \\ 
    \texttt{\small \{wuxiaoshi@link, hsli@ee\}.cuhk.edu.hk},~\texttt{\small \{zhufeng, zhaorui\}@sensetime.com}
}

\maketitle

\begin{abstract}
   Open-vocabulary detection (OVD) is an object detection task aiming at detecting objects from novel categories beyond the base categories on which the detector is trained.
   Recent OVD methods rely on large-scale visual-language pre-trained models, such as CLIP,  for recognizing novel objects.
   We identify the two core obstacles that need to be tackled when incorporating these models into detector training:
   (1) the distribution mismatch that happens when applying a VL-model trained on whole images to region recognition tasks;
   (2) the difficulty of localizing objects of unseen classes.
   To overcome these obstacles, we propose CORA, a DETR-style framework that adapts \textbf{C}LIP for \textbf{O}pen-vocabulary detection by \textbf{R}egion prompting and \textbf{A}nchor pre-matching.
   Region prompting mitigates the whole-to-region distribution gap by prompting the region features of the CLIP-based region classifier.
   Anchor pre-matching helps learning generalizable object localization by a class-aware matching mechanism.   
   We evaluate CORA on the COCO OVD benchmark, where we achieve 41.7 AP50 on novel classes, which outperforms the previous SOTA by 2.4 AP50 even without resorting to extra training data.
   When extra training data is available, we train CORA$^+$ on both ground-truth base-category annotations and additional pseudo bounding box labels computed by CORA.  
   CORA$^+$ achieves 43.1 AP50 on the COCO OVD benchmark and 28.1 box APr on the LVIS OVD benchmark.
   The code is available at \href{https://github.com/tgxs002/CORA}{https://github.com/tgxs002/CORA}.
\end{abstract}

\section{Introduction}
\label{sec:introduction}


Object detection is a fundamental vision problem that involves localizing and classifying objects from images.
Classical object detection requires detecting objects from a closed set of categories. 
Extra annotations and training are required if objects of unseen categories need to be detected. 
It has attracted much attention on detecting novel categories without tedious annotations, or even detect object from new category, which is currently referred as open-vocabulary detection (OVD)~\cite{OVRCNN}.

Recent advances on large-scale vision-language pre-trained models, such as CLIP~\cite{clip}, enable new solutions for tackling OVD.
CLIP learns a joint embedding space of images and text from a large-scale image-text dataset, which shows remarkable capability on visual recognition tasks. 
The general idea of applying CLIP for OVD is to treat it as an open-vocabulary classifier.
However, there are two obstacles hindering the effective use of CLIP on tackling OVD.

\noindent\textbf{How to adapt CLIP for region-level tasks?}
One trivial solution is to crop regions and treat them as separate images, which has been adopted by multiple recent works ~\cite{vild, OVDETR, medet, bridging}. 
But the distribution gap between region crops and full images leads to inferior classification accuracy. 
MEDet~\cite{medet} mitigates this issue by augmenting the text feature with image features.
However, it requires extra image-text pairs to prevent overfitting to so-called ``base'' classes that are seen during training.
RegionCLIP ~\cite{regionclip} directly acquires regional features by RoIAlign~\cite{maskrcnn}, which is more efficient but cannot generalize well to novel classes without finetuning. The finetuning is costly when adopting a larger CLIP model.

\noindent\textbf{How to learn generalizable object proposals?} 
ViLD~\cite{vild}, OV-DETR~\cite{OVDETR}, Object-Centric-OVD~\cite{bridging}, RegionCLIP~\cite{regionclip} need RPN or class-agnostic object detectors~\cite{mvit} to mine potential novel class objects.
However, these RPNs are strongly biased towards the base classes on which they are trained, while perform poorly on the novel classes.
MEDet~\cite{medet} and VL-PLM~\cite{vlplm} identify this problem and adopt several handcrafted policies to rule out or merge low-quality boxes, but the performance is still bounded by the frozen RPN. 
OV-DETR~\cite{OVDETR} learns generalizable object localization by conditioning box regression on class name embeddings, but at the cost of efficiency issue induced by repetitive per-class inference.

In this work, we propose a new framework based on DEtection TRansformers (DETR)~\cite{DETR} that incorporates CLIP into detector training to achieve open-vocabulary detection without additional image-text data.
Specifically, we use a DETR-style object localizer for class-aware object localization, and the predicted boxes are encoded by pooling the intermediate feature map of the CLIP image encoder, which are classified by the CLIP text encoder with class names.
However, there is a distribution gap between whole-image features from CLIP's original visual encoder and the newly pooled region features, leading to an inferior classification accuracy.
Thus, we propose Region Prompting to adapt the CLIP image encoder, which boosts the classification performance, and also demonstrates better generalization capability than existing methods.
We adopt DAB-DETR~\cite{dabdetr} as the localizer, in which object queries are associated with dynamic anchor boxes. 
By pre-matching the dynamic anchor boxes with the input categories before box regression (Anchor Pre-Matching), class-aware regression can be achieved without the cost of repetitive per-class inference.

We validate our method on COCO~\cite{COCO} and LVIS v1.0~\cite{LVIS} OVD benchmarks.
On the COCO OVD benchmark, our method improves AP50 of novel categories over the previous best method~\cite{regionclip} by 2.4 AP50 without training on extra data, and achieves consistent gain on CLIP models of different scales.
When compared under a fairer setting with extra training data, our method significantly outperforms the existing methods by 3.8 AP50 on novel categories and achieves comparable performance on base categories.
On the LVIS OVD benchmark, our method achieves 22.2/28.1 APr with/w.o. extra data, which significantly outperforms existing methods that are also trained with/w.o. extra data.
By applying region prompting on the base classes of COCO, the classification performance on the novel classes is boosted from 63.9\% to 74.1\%, whereas other prompting or adaptation methods easily bias towards the base classes.

The contributions of this work are summarized as follows:
(1) Our proposed region prompting effectively mitigates the gap between whole image features and region features, and generalize well in the open-vocabulary setting.
(2) Anchor Pre-Matching enables DETR for generalizable object localization efficiently.
(3) We achieve state-of-the-art performance on COCO and LVIS OVD benchmarks.

\section{Related Works}
\label{sec:related}
\noindent\textbf{Open-Vocabulary Object Detection}
OVR-CNN~\cite{OVRCNN} firstly put forth this new formulation of detection, and proposes its baseline solution by aligning region features with nouns in captions that are paired with the image. 
Mingfeng~\cite{mingfeng} \etal mine pseudo labels by utilizing the localization ability of pre-trained vision-language models. 
PromptDet~\cite{promptdet} addresses the gap between image and region classification by adding learnable prompts when encoding the class names, namely regional prompt learning (RPL), which is expected to generalize from base to novel categories. 
OV-DETR~\cite{OVDETR} is the first DETR-style open-vocabulary detector, which proposes conditional matching to solve the missing novel class problem in assignment, but at the cost of inefficient inference.
RegionCLIP~\cite{regionclip} propose a second-stage pre-training mechanism to adapt the CLIP model to encode region features, and demonstrates its capability on OVD and zero-shot transfer setting.
GLIP~\cite{glip} jointly learns object localization and VL alignment. Matthias \etal~\cite{simple} proposes to finetune a VL aligned model for detection, while we fix the pre-trained VL model for better generalization towards novel categories.

\noindent\textbf{Detection Transformers}
DETR~\cite{DETR} is an object detection architecture based on transformers that formulates object detection as a set-to-set matching problem, which greatly simplifies the pipeline. 
Several works address the slow convergence problem of DETR by architectural improvement~\cite{deformable, SMCA, dabdetr, SAMDETR} or special training strategies~\cite{groupdetr, hybridmatching}. 
Zhu \etal~\cite{deformable} proposes multi-scale deformable attention module to efficiently aggregate information from multi-scale feature maps.
Gao \etal~\cite{SMCA} proposes to modulate the cross-attention in the transformer decoder by anchor box coordinates to accelerate the detector convergence.
DAB-DETR~\cite{dabdetr} formulates the queries in DETR architecture as anchor boxes, which accelerates detector training.
Chen \etal~\cite{SAMDETR} proposes Group DETR, which adds auxiliary object queries during training to take advantage of one-to-many matching for faster convergence.

\noindent\textbf{Prompt Tuning}
Prompting is originated from NLP, and it refers to prepending task instructions before the input sequence to give the language model the hint about the task~\cite{gpt3}. 
Later works ~\cite{prefixtuning, ptuningv2} explores tuning continuous prompt vectors when few-shot data is available.
VPT~\cite{vpt}, Visual Prompting~\cite{VisualPM, bahng2022visual} explore prompting in the pixel space. 
~\cite{video1} and ~\cite{video2} prompts pre-trained model for video recognition tasks. 
~\cite{classaware} proposes class-aware visual prompt tuning to generalize the learned prompts to unseen categories.
Recent works demonstrate that prompt tuning is an effective and parameter-efficient way to adapt large-scale pre-trained models to downstream tasks.

\begin{figure*}
  \centering
    \includegraphics[width=0.95\linewidth]{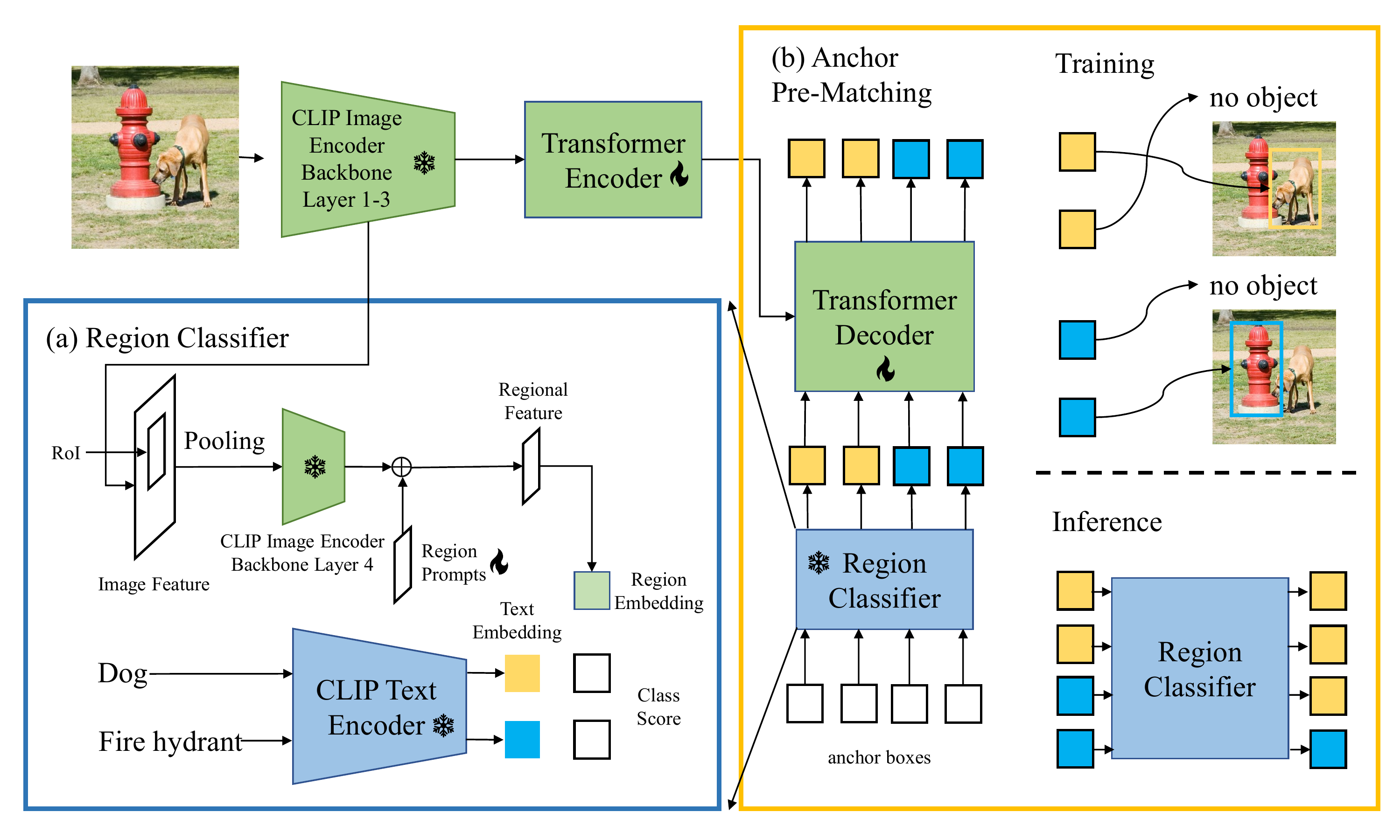}
  \caption{Overview of our method. The image is encoded into a feature map by the CLIP image encoder for both localization and classification. The regional feature is extracted by pooling the feature map, and then prompted before classified by the CLIP class name embeddings. The anchor boxes are pre-matched and conditioned on a class before decoding. During training, per-class post-matching is conducted. During inference, the box predictions are classified by the region classifier. }
  \label{fig:overview}
\end{figure*}


\section{Method}
\label{sec:method}


Open-vocabulary detection (OVD) is an object detection task aiming at detecting objects from novel categories beyond the base categories on which the detector is trained. Formally, the detector is trained on a detection dataset with base-category $C^B$ box annotations, and tested on a new input image $I \in \mathbb{R}^{H\times W\times 3}$ to detect objects belonging to a novel category set $C^N$, where $C^B\cap C^N=\emptyset $. 
In this section, we introduce CORA, a framework that adapts CLIP for the OVD task by Region prompting and Anchor pre-matching.
For fairer comparisons with existing methods, we also experiment with a broader setting where extra data is available, which is referred to as CORA$^+$, and will be introduced along with the experiments.


\subsection{Overview}

\textcolor{black}{The overall framework of CORA is illustrated in \cref{fig:overview}.}
Given an image as input, we acquire the spatial feature map using the ResNet backbone from the pre-trained CLIP image encoder, which is shared by both region classification and object localization branches.
Unlike conventional detectors, localization and classification are decoupled and sequentially conducted in our framework to better fit the characteristic of the OVD problem.
We train a DETR-style object localizer that refines a set of object queries together with their associated anchor boxes to localize the objects, which are then classified by a region classifier adapted from CLIP.

\noindent\textbf{Region Classification.}
Given a region to be classified (anchor box or box prediction), we adopt RoIAlign to obtain the region feature, followed by the attention pooling module of CLIP to generate region embeddings, which can be classified by class embeddings obtained from the CLIP text encoder, as done in CLIP.
We name this module as CLIP-based region classifier (\cref{fig:overview}-(a)).

\noindent\textbf{Object Localization.}
The visual feature map is firstly refined by the DETR-like encoder, and then fed into the DETR-like decoder.
The queries of anchor boxes are firstly classified by the CLIP-based region classifier, which are then conditioned on their predicted labels before iteratively refined by the DETR-like decoder for better localization. The decoder also estimates the matchability of the query with the previously predicted label.
During training, the predicted boxes are one-to-one matched with ground truth boxes that has the same labels (\cref{fig:overview}-(b)), and trained as in DETR.
In the inference stage, the class labels of the boxes are adjusted by the CLIP-based region classifier.

As mentioned in \cref{sec:introduction}, there are two obstacles to be addressed:
(1) object detection conducts recognition on image regions, while the CLIP model is trained on whole-image input, leading to a distribution gap that hinders classification performance.
(2) the detector needs to learn object localization for novel classes, while we only have annotations on a limited number of base classes. 
To solve the first obstacle, we propose region prompting to modulate the region features for better generalizable region embeddings, which will be introduced in \cref{sec:region_prompting}.
To solve the second obstacle, we put forward anchor pre-matching to encourage class-aware object localization that can generalize to novel classes during inference, which will be introduced in \cref{sec:proposal_pre_matching}.

\begin{figure}[t]
  \centering
  \begin{subfigure}{1.0\linewidth}
  \centering
    \includegraphics[width=1.\linewidth]{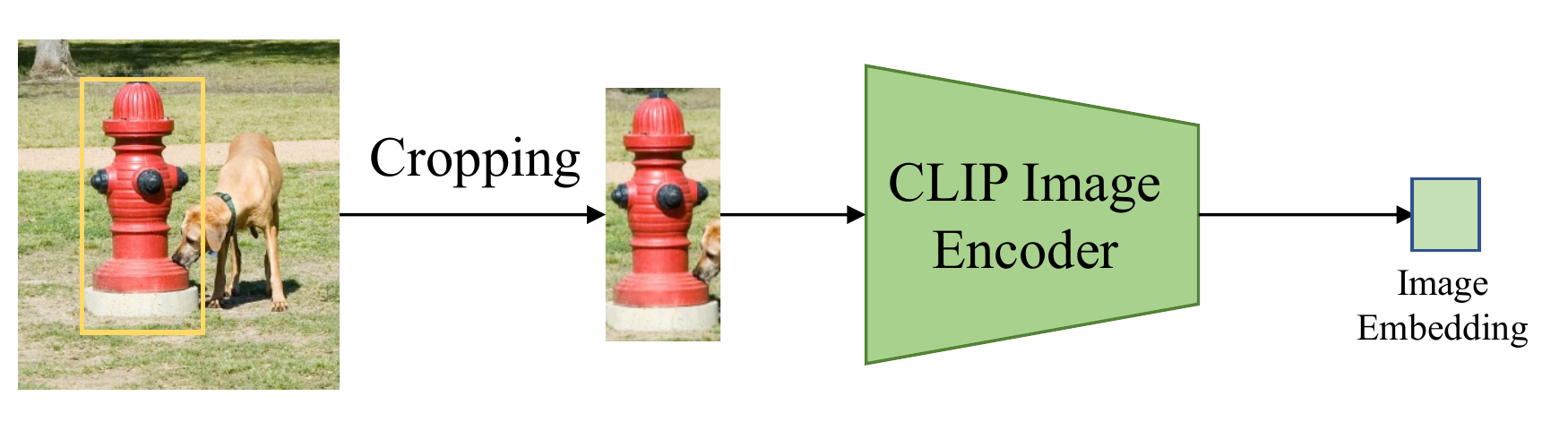}
    \caption{Whole-image encoding}
    \label{fig:whole_image}
  \end{subfigure}
  \hfill
  \begin{subfigure}{1.0\linewidth}
  \centering
    \includegraphics[width=1.\linewidth]{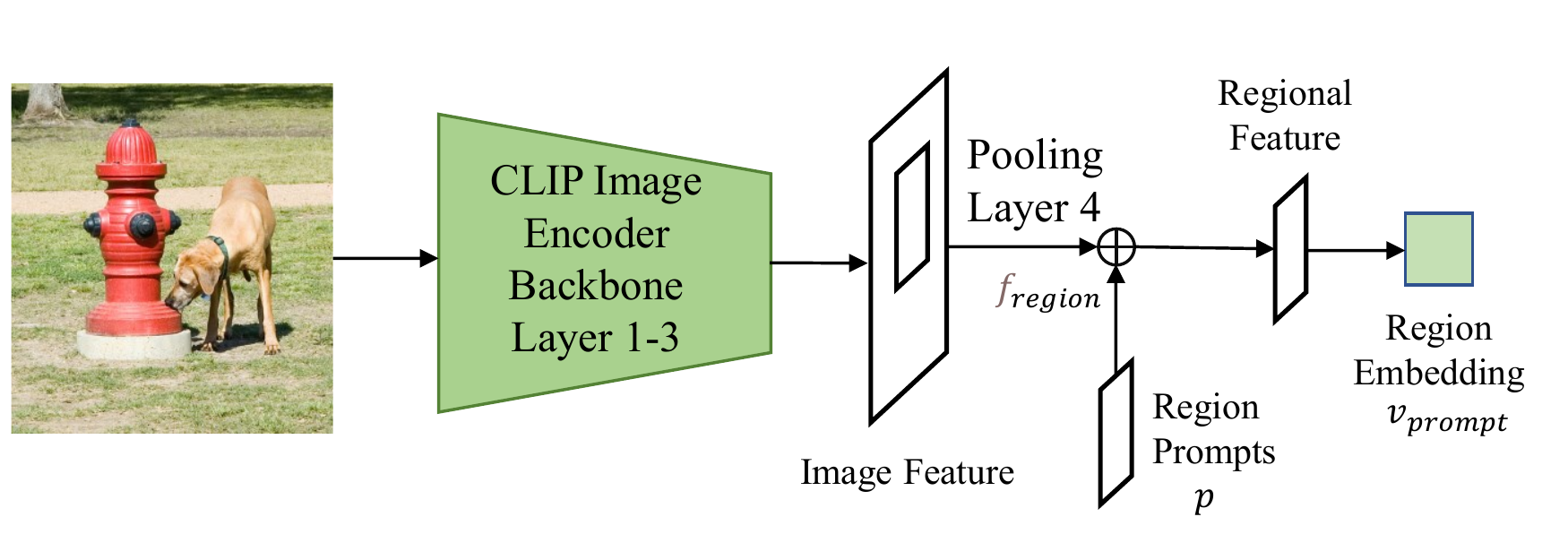}
    \caption{Regional encoding}
    \label{fig:regional}
  \end{subfigure}
   \caption{Comparison between the CLIP-based region classifier and the vanilla pipeline. We pool regional feature from the feature map, rather than cropping the region patch and classify them as a separate image.}
   \label{fig:region}
\end{figure}

\subsection{Region Prompting}
\label{sec:region_prompting}
OVD requires the detector to classify image regions into a given category list.
In this section, we will elaborate how a pre-trained CLIP model is adapted to formulate the CLIP-based region classifier.
Given the CLIP model, region classification can be realized by comparing the similarity between the regional embedding from the CLIP image encoder and the class name embedding from the CLIP text encoder.

\noindent\textbf{Region Prompting.}
As illustrated in \cref{fig:region}, given an image and a set of region of interest (RoI), we firstly encode the whole image into a feature map by the CLIP encoder's first 3 blocks, which are then pooled by RoIAlign~\cite{fasterrcnn} either according to anchor boxes or predicted boxes into region features, before encoded by the last block of the CLIP image encoder backbone.
There exists a distribution gap between the CLIP image encoder's whole-image feature map and the pooled regional features.
We propose region prompting to fix the misalignment by augmenting the region feature with learnable prompts $p \in \mathbb{R}^{S\times S\times C}$, where $S$ is the spatial size of the regional feature, and $C$ is the dimension of the regional feature.
Specifically, given the input regional feature $f_\mathrm{region}$, the region prompting is conducted as
\begin{equation}
  v_\mathrm{prompt} = P(f_\mathrm{region} \oplus p),
  \label{eq:logits}
\end{equation}
where $\oplus$ denotes element-wise addition, $P$ is the attention pooling module of the CLIP visual encoder.

\noindent\textbf{Optimizing Region Prompts.}
We train the region prompts on a detection dataset with base-class annotations.
The class name embeddings are pre-computed by the CLIP text encoder, which are used as classifier weights later.
We train the prompts by a standard cross-entropy loss to classify the ground truth boxes with their pooled regional features $f_\mathrm{region}$.
When optimizing the region prompts, we keep other model weights frozen, and only make the region prompts to be learned. 

\noindent\textbf{Comparison with existing methods.}
The previous common practice for region classification with CLIP is to crop the RoIs, and encode them as separate images, before comparing with text embeddings.
This pipeline is not efficient when encoding regions with overlaps, since the overlapping regions are encoded more than once in different region crops.
Its accuracy also suffers from the missing context information.
In contrast, our regional prompting is more efficient and preserves richer context.

The region prompts contain less than 1M parameters, which is in line with recent advances of the prompt tuning and adapter literatures.
Region prompting generalizes well to unseen novel classes.
We attribute the generalization capability to the fact that region prompting directly \textbf{fix} the distribution mismatch right after where it occurs (after region pooling), whereas the existing methods tweak irrelevant parameters to \textbf{compensate} for the distribution mismatch.


\begin{figure}[t]
  \centering
  \begin{subfigure}{1.0\linewidth}
    \includegraphics[width=1.\linewidth]{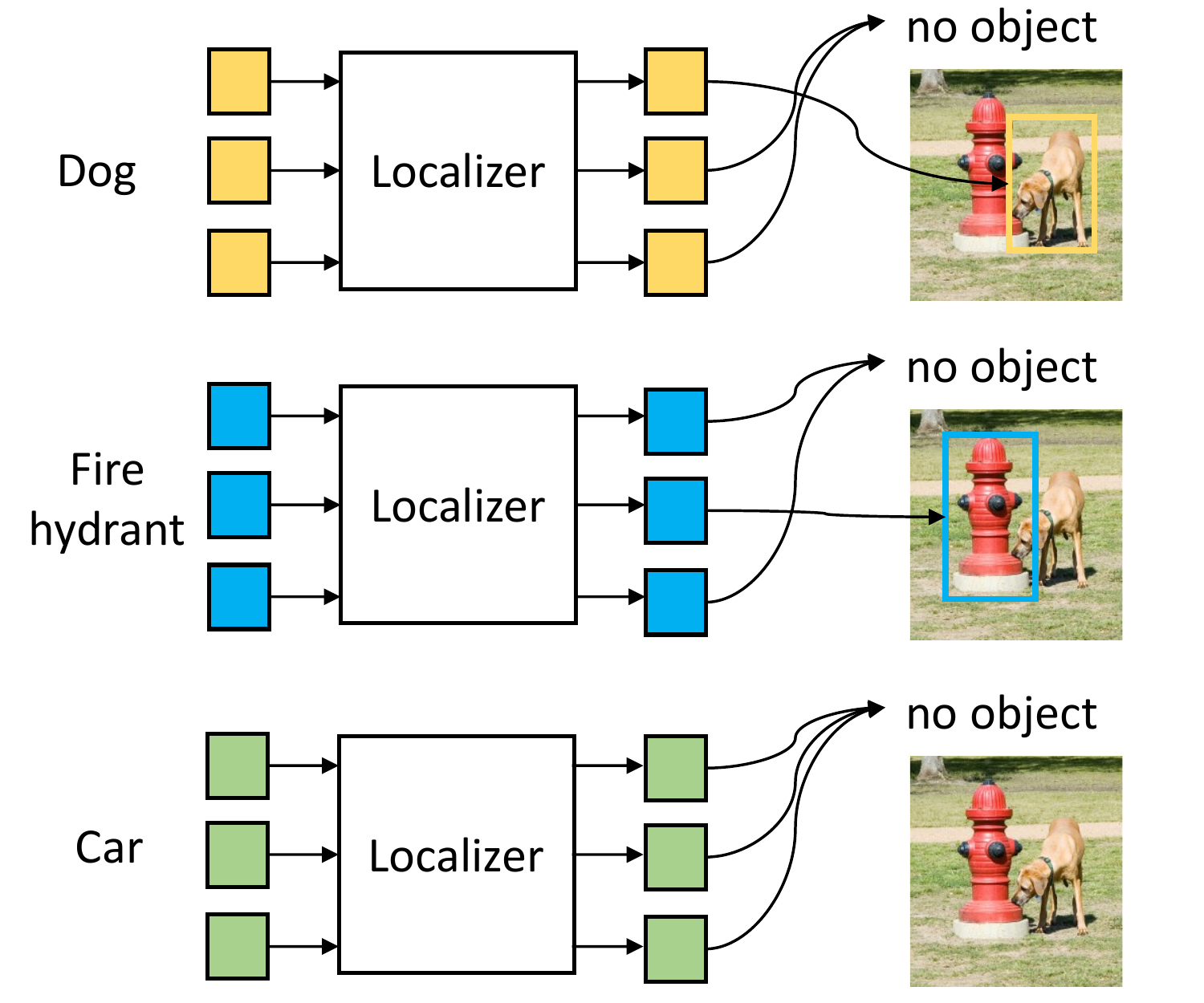}
    \caption{Conditional matching}
    \label{fig:cond_matching}
  \end{subfigure}
  \hfill
  \begin{subfigure}{0.9\linewidth}
    \includegraphics[width=1.\linewidth]{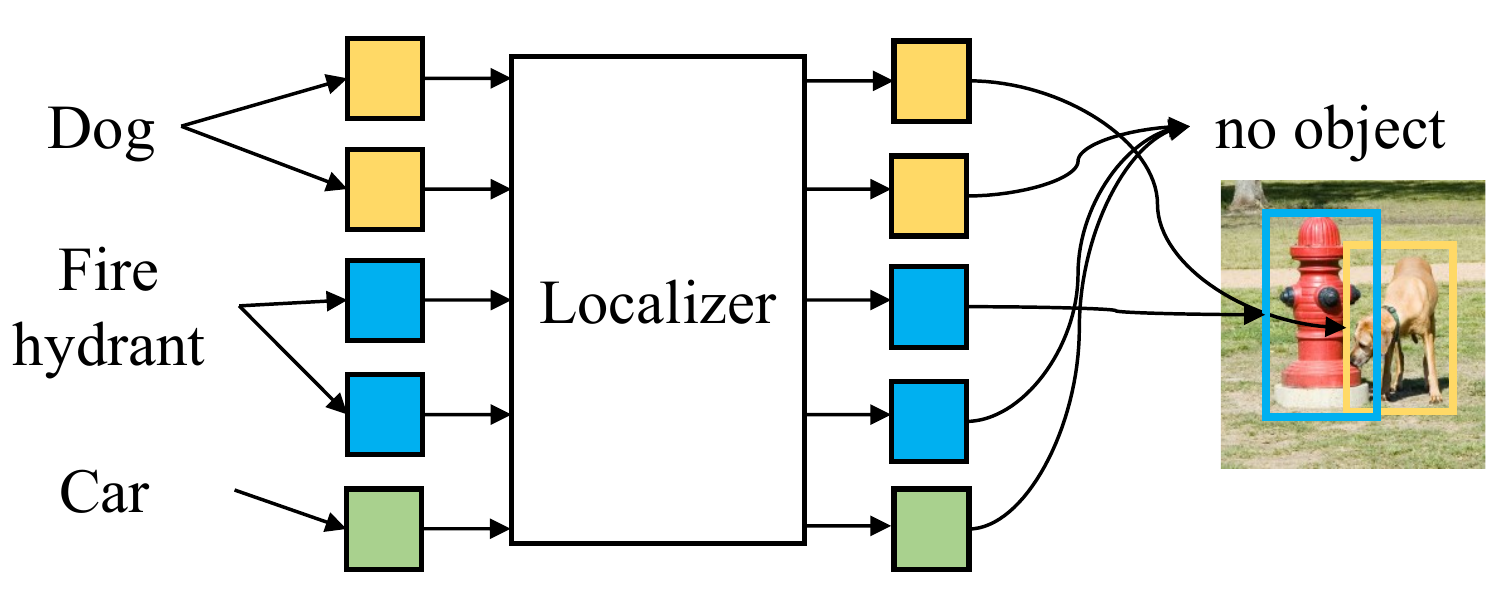}
    \caption{Anchor pre-matching}
    \label{fig:anchor_prematching}
  \end{subfigure}
   \caption{Comparison between anchor pre-matching and conditional matching. Anchor pre-matching decodes a constant number of object queries, and assigns different numbers of object queries based on the image content, avoiding repetitive decoding as in conditional matching.}
   \label{fig:ppm}
\end{figure}

\subsection{Anchor Pre-Matching}
\label{sec:proposal_pre_matching}

Region prompting helps solve the region classification problem. Object localization is the other critical sub-task of object detection. 
Considering the inferior performance of pre-trained RPN on novel classes, we introduce a class-aware query-based object localizer, which demonstrates better generalization capability on unseen classes.
As shown in \cref{fig:overview}, Given the visual feature map from the frozen CLIP image encoder, the object queries are pre-matched to the class name embeddings by the CLIP text encoder.


\noindent\textbf{Anchor Pre-Matching. } The object localizer is implemented by a DETR-style transformer encoder-decoder structure, where the encoder refines the feature map, and the decoder decodes a set of object queries into box predictions. We adopt DAB-DETR~\cite{dabdetr}, where each object query is associated with an anchor box. Each ground truth box is pre-matched to a set of queries with the same label. The label $\hat{c}_i$ of an object query is assigned by classifying the associated anchor box $b_i$
\begin{equation}
  \hat{c}_i = \underset{c \in C^B}{\arg\max} \ \mathrm{cosine}(v_i, l_c),
  \label{eq:prematch}
\end{equation}
where $v_i$ is the region feature of anchor box $b_i$, $l_c$ is the class name embedding of class $c$, and $\mathrm{cosine}$ denotes the cosine similarity.
After pre-matching, each object query conditions on the predicted class embedding to allow class-aware box regression. 
The conditioned object query is given by
\begin{equation}
  q_i = \mathrm{MLP}(l_\mathrm{c_i}).
  \label{eq:query}
\end{equation}
The DETR-like decoder iteratively refines each object query with its associated anchor box $(q_i, b_i)$ into $\hat{y_i} = (\hat{p}_i, \hat{b}_i)$, where $\hat{b}_i$ is the refined box coordinates and $\hat{p}_i$ is the matching probability to the query's pre-matched class $\hat{c}_i$.

Given the model predictions, the assignment between ground truth boxes and model predictions is conducted by performing bipartite matching for each class separately.
We only allow each ground truth box to be assigned to the prediction with the same pre-matched label, in order to enforce the decoder to be aware of the conditioned text embedding.

Specifically, for class $c$, given the $N^c$ box predictions $\hat{y}^c = \{\hat{y}_i\ |\ \hat{c}_i=c\}$ that are pre-matched to class $c$, and the set of ground truth boxes $y^c$ in class $c$, we optimize a permutation of $N^c$ elements $\sigma \in \mathfrak{S}_\mathrm{N^c}$ that minimizes the following cost
\begin{equation}
  \hat{\sigma}_c = \underset{\sigma \in \mathfrak{S}_\mathrm{N^c}}{\arg\max} \sum_i^{N^c} \mathcal{L}_\mathrm{cost}(y_i^c, \hat{y}^c_\mathrm{\sigma(i)}),
  \label{eq:matching}
\end{equation}
where the matching cost is defined as
\begin{equation}
  \mathcal{L}_\mathrm{cost}(y, \hat{y}) = \mathcal{L}_\mathrm{match}(p,\hat{p}) + \mathcal{L}_\mathrm{box}(b,\hat{b}).
  \label{eq:cost}
\end{equation}
$\mathcal{L}_\mathrm{match}(p,\hat{p})$ is a binary classification loss, and $\mathcal{L}_\mathrm{box}(b,\hat{b})$ characterizes the localization error of $\hat{b}$ w.r.t $b$. 
In our case, we implement $\mathcal{L}_\mathrm{match}$ by the focal loss ~\cite{focal}. $\mathcal{L}_\mathrm{box}$ is implemented by a weighted sum of $L_1$ loss and GIoU~\cite{giou} loss following prior works.

The model is optimized by the following loss
\begin{equation}
\begin{split}
  \mathcal{L} & = \sum_{c\in C^B} 
  \mathcal{L}_\mathrm{match}(p^c,\hat{p}^c_\mathrm{\hat{\sigma}_c}) + \mathcal{L}_\mathrm{box}(b^c,\hat{b}^c_\mathrm{\hat{\sigma}_c}) \\
  & = \lambda_\mathrm{focal} \mathcal{L}_\mathrm{focal} + \lambda_\mathrm{L_1} \mathcal{L}_\mathrm{L_1} + 
  \lambda_\mathrm{GIoU} \mathcal{L}_\mathrm{GIoU}.
\end{split}
  \label{eq:loss}
\end{equation}

During inference, we adopt the region classifier introduced in \cref{sec:region_prompting} to classify the predicted boxes $\{\hat{b}_i\}$ for better classification accuracy. The class score is multiplied by the pre-matching score to account for the box quality
\begin{equation}
  \mathrm{P}(\hat{b}_i \in c) = \hat{p}_i \mathrm{cosine}(\hat{v}_i, l_c).
  \label{eq:prematch}
\end{equation}

\noindent\textbf{Comparison with Conditional Matching~\cite{OVDETR}.} 
Conditional Matching in OV-DETR~\cite{OVDETR} also proposes to condition the queries on the text embedding for class-aware regression.
But it suffers from repetitive per-class inference.
Specifically, as shown in \cref{fig:ppm}, each class in $C^N$ needs to be separately localized with the same group of query, which means both computation and memory consumption scales linearly with the vocabulary size.
During training, the number of negative classes sampled in each iteration is limited due to the memory constraint, which hinders convergence. 
During inference, repetitive per-class decoding is required and results in low inference efficiency, especially when the vocabulary size is large.

Contrary to conditional matching, our anchor pre-matching mechanism assigns anchor boxes for different classes adaptively according to the image content, which ensures a constant number of query decoupled from the category size. 
By anchor pre-matching, all the classes can be decoded together in one pass, eliminating the need for repetitive per-class decoding.

To improve the generalization capability and training convergence of open-vocabulary detectors equipped with Anchor Pre-Matching, we also introduce two effective training techniques, namely, ``Drop Class'' and ``CLIP-Aligned Labeling''.

\noindent\textbf{Class Dropout.} The generalization capability of the model can be further boosted by randomly dropping categories during training.
Since our goal is to train a detector that can detect objects from a user specified category list, training on a fixed list of categories leads to bias. 
We mitigate this bias by randomly dropping out the base categories during training.
For efficiency reason, we implement this idea by splitting the base classes into two complementary groups and train on both of them, instead of training on one group while dropping the other group.
Specifically, in each training iteration, we split $C^B$ and the ground truth boxes at a probability of $p$, and train the detector on the same image with two complementary sets of categories. 
It enforces the model to condition its prediction on the query categories.
Since a ground truth box in one group does not appear in the other group,
the model needs to be aware of the categories to treat differently for the two sets of annotations.

\noindent\textbf{CLIP-Aligned Labeling.} Directly training the localizer on the original COCO dataset suffers from convergence issue.
By anchor pre-matching mechanism, a ground truth box incorporates training only when at least one query with the same pre-matched label exists.
Otherwise, it is ignored, which hinders convergence. 
This issue can be partially attributed to the inaccurate anchor box.
However, even if a ground truth box has an accurate anchor box, it may still be ignored due to the limited recognition accuracy limitation of the region classifier, or in other words, the ground truth box label is not aligned with the CLIP region classifier used for pre-matching.
Thus, we relabel the boxes in the training dataset with the region classifier, which we refer to as CLIP-Aligned labeling. With this technique, more ground truth boxes can be matched.

\begin{table*} [t]
    \centering
\resizebox{0.9\textwidth}{!}{

\begin{tabular}{l|llc|ccc}
\toprule
\multicolumn{1}{c|}{}                         & \multicolumn{3}{c|}{Detector Training}                                                        & \multicolumn{3}{c}{Generalized (17 + 48)}                                 \\
\multicolumn{1}{c|}{\multirow{-2}{*}{Method}} & \multicolumn{1}{c}{Extra Dataset} & \multicolumn{1}{c}{Pre-train Model} & Require Novel Class & Novel         & {\color[HTML]{C0C0C0} Base} & {\color[HTML]{C0C0C0} All}  \\
\midrule
OVR-CNN~\cite{OVRCNN}                                       & COCO Captions~\cite{cococaptions}                     & -                                   & \xmark              & 22.8          & {\color[HTML]{C0C0C0} 46.0} & {\color[HTML]{C0C0C0} 39.9} \\
Detic~\cite{detic}                                         & COCO Captions~\cite{cococaptions}                     & CLIP (text encoder)                 & \xmark              & 27.8          & {\color[HTML]{C0C0C0} 47.1} & {\color[HTML]{C0C0C0} 45.0} \\
\midrule
RegionCLIP~\cite{regionclip}                                    & CC3M~\cite{cc3m}                              & CLIP (RN50)                         & \xmark              & 31.4          & {\color[HTML]{C0C0C0} 57.1} & {\color[HTML]{C0C0C0} 50.4} \\
VL-PLM~\cite{vlplm}                                        & -                                 & CLIP (RN50)                         & \cmark              & 34.4          & {\color[HTML]{C0C0C0} 60.2} & {\color[HTML]{C0C0C0} 53.5} \\
CORA (Ours)                                   & -                                 & CLIP (RN50)                         & \xmark              & \textbf{35.1} & {\color[HTML]{C0C0C0} 35.5} & {\color[HTML]{C0C0C0} 35.4} \\
\midrule
ViLD~\cite{vild}                                          & -                                 & CLIP (ViT-B/32)                     & \cmark              & 27.6          & {\color[HTML]{C0C0C0} 59.9} & {\color[HTML]{C0C0C0} 51.3} \\
OV-DETR~\cite{OVDETR}                                       & -                                 & CLIP (ViT-B/32)                     & \cmark              & 29.4          & {\color[HTML]{C0C0C0} 61.0} & {\color[HTML]{C0C0C0} 52.7} \\
MEDet~\cite{medet}                                         & COCO Captions~\cite{cococaptions}                     & CLIP (ViT-B/32)                     & \xmark              & 32.6          & {\color[HTML]{C0C0C0} 54.0} & {\color[HTML]{C0C0C0} 49.4} \\
\midrule
RegionCLIP~\cite{regionclip}                                    & CC3M~\cite{cc3m}                              & CLIP (RN50x4)                       & \xmark              & 39.3          & {\color[HTML]{C0C0C0} 61.6} & {\color[HTML]{C0C0C0} 55.7} \\
CORA (Ours)                                   & -                                 & CLIP (RN50x4)                       & \xmark              & 41.7 & {\color[HTML]{C0C0C0} 44.5} & {\color[HTML]{C0C0C0} 43.8} \\
CORA$^+$ (Ours)                                   & COCO Captions~\cite{cococaptions}                   & CLIP (RN50x4)     & \xmark              & \textbf{43.1} & {\color[HTML]{C0C0C0} 60.9} & {\color[HTML]{C0C0C0} 56.2} \\
\bottomrule
\end{tabular}
}
\caption{Main results on the COCO OVD benchmark. We report AP50 as the evaluation metric. The baseline methods are grouped by their pre-trained model. We also list the extra dataset requirement of each method, and whether they require the novel class to be provided during training.}
\label{tab:ovd_main}
\end{table*}
\begin{table}[]
\centering
\resizebox{.47\textwidth}{!}{
\begin{tabular}{l|ll|c}
\toprule
\multicolumn{1}{c|}{\multirow{2}{*}{Method}} & \multicolumn{2}{c|}{Detector Training}                               & \multicolumn{1}{l}{LVIS} \\
\multicolumn{1}{c|}{}                        & \multicolumn{1}{c}{Ext. Data} & \multicolumn{1}{c|}{Pre-train Model} & APr                      \\
\midrule
ViLD~\cite{vild}                                         & -                             & CLIP (ViT-B/32)                      & 16.3                     \\
OV-DETR~\cite{OVDETR}                                      & -                             & CLIP (ViT-B/32)                      & 17.4                     \\
\midrule
RegionCLIP~\cite{regionclip}
& CC3M                          & CLIP (RN50x4)                        & 22.0                     \\
CORA (Ours)                                  & -                             & CLIP (RN50x4)                        & 22.2                       \\
MEDet~\cite{medet}                                        & CC3M                          & CLIP (ViT-B/32)                      & 22.4                     \\
Detic~\cite{detic}                                        & IN-21k                  & CLIP (text encoder)                  & 26.2                     \\
CORA$^+$ (Ours)                                  & IN-21k                             & CLIP (text encoder)                        & \textbf{28.1}                       \\
\bottomrule
\end{tabular}
}
\caption{Results on the LVIS~\cite{LVIS} OVD benchmark.}
\label{tab:ovd_lvis}
\end{table}

\section{Experiments}
\label{sec:exp}

\textcolor{black}{In this section, we comprehensively evaluate our CORA on the open-vocabulary detection task. Datasets and evaluation protocols are introduced in Sec.~\ref{sec:dataset}, and implementation details of our method are provided in Sec.~\ref{sec:implementation}. We compare with state-of-the-art methods in Sec.~\ref{sec:sotas}, demonstrating advantages of our CORA, and then validate the effectiveness of the proposed Region Prompting and Anchor Pre-Matching in Sec.~\ref{sec:region_prompt} and Sec.~\ref{sec:anchor_pre-matching}, respectively.}

\subsection{Dataset \& Training \& Evaluation}
\label{sec:dataset}

Following the convention of the COCO OVD benchmark proposed in ~\cite{zsd}, the 80 classes in the COCO dataset~\cite{COCO} classes are divided into 48 base classes and 17 novel classes.
The model is trained on the 48 base classes, which contains 107,761 images and 665,387 instances.
The model is then evaluated on the validation set of novel classes, which contains 4,836 images and 33,152 instances from both the 48 base classes and the 17 novel classes.
We also conduct experiments on the LVIS v1.0~\cite{LVIS} dataset.
On LVIS dataset, the model is trained on 461 common classes and 405 frequent classes, which contains 100,170 images and 1,264,883 instances.
After training, the model is evaluated on the LVIS validation set, which contains 19,809 images and 244,707 instances.

To make fairer comparisons with other methods, we propose CORA$^+$, which utilizes extra dataset or target novel class names.
CORA$^+$ is a detector trained with both ground-truth base-category annotations and additional pseudo bounding box labels. When target class names are provided, we generate pseudo boxes of novel classes on the base training dataset using CORA, and when extra image-text dataset (text could be captions or class names) is available, 
pseudo boxes of all objects mentioned by the corresponding text are generated.
We use standard detector architecture and training target to train CORA$^+$.
SAM-DETR~\cite{SAMDETR} is used for the experiment on COCO and CenterNet2~\cite{centernet2} is used for the experiment on LVIS.

In the OVD task, we evaluate our model under the ``generalized'' setting, in which the model needs to predict objects from both base and novel classes, and then evaluated on novel objects.
On the COCO benchmark, we take AP50 as our evaluation metric, which counts the average precision at an intersection over union (IoU) of 50\% for each class, and then averages among all the classes.
For the LVIS OVD benchmark, we evaluate on the full validation dataset, and report the mean AP of boxes from the novel classes to compare with prior works~\cite{regionclip}. 
For the region classification task, we let the model classify the ground truth boxes in the COCO dataset, the performance is evaluated in terms of mAP.

\subsection{Implementation Details}
\label{sec:implementation}

\noindent\textbf{Model Specifications.} We use DAB-DETR~\cite{dabdetr} as the object localizer.
Specifically, localizer is configured to have 1,000 object queries, 3 encoder layers and 6 decoder layers.
We use a multi-layer perceptron (MLP) with 128 hidden neurons to transform class name embeddings into an object query. 
Following CLIP~\cite{clip}, each class embedding is computed as the average text embedding of the class name over 80 context prompts.
In class dropout, each class is randomly assigned to one of the two groups with equal probabilities.
When adopting region prompting on LVIS, classes in the common and frequent group are sampled with equal weights, meaning that the objects in the less frequent classes are over sampled.
When training our method on LVIS, we sample 100 categories (including the ground truth categories) in each iteration.
Since the number of classes in the LVIS dataset is much larger than that of COCO, we relax the matching constraint in anchor pre-matching, such that ground truth boxes can be post-matched to the anchor boxes with a similar label.
Specifically, classes with a cosine similarity greater than 0.7 are considered similar.

\noindent\textbf{Training \& Hyperparameters.}
We train the region prompts for 5 epochs with a base learning rate of $10^{-4}$, which decays after the $4^{th}$ epoch by a factor of 0.1. 
The localizer is trained for 35 epochs with a learning rate of $10^{-4}$ without learning rate decay. 
Both the region prompts and the localizer are trained with batch size 32 by the AdamW optimizer~\cite{adamw} with $10^{-4}$ weight decay. 
We apply gradient clipping with a maximal norm of 0.1. 
To stabilize training, we evaluate on the exponential moving average (EMA) of the model after training. 
The class dropout probability $p$ is set as 0.2. 
$\lambda_\mathrm{focal}$, $\lambda_\mathrm{L_1}$ and $\lambda_\mathrm{GIoU}$ are set as 2.0, 5.0, 2.0,  respectively.
For the experiment on LVIS, we use repeat factor sampling~\cite{LVIS} with default hyperparameters to balance the training samples.
We use non-maximum suppression (NMS) with an IoU threshold 0.5 during inference.  

\subsection{\textcolor{black}{Comparison with State-of-the-Art Methods}}
\label{sec:sotas}

\cref{tab:ovd_main} summarizes our main results. 
Since the pre-trained model is crucial to the open-vocabulary capability of the detector, we compare our method with baseline methods that are trained with the same CLIP model.
When compared with methods trained on CLIP RN50, CORA outperforms VL-PLM by 0.7 AP50 on novel classes.
With a larger pre-trained model, our method improves the previous state-of-the-art RegionCLIP~\cite{regionclip} by 2.4 AP50.
When extra data is available, the performance can be further boosted to 43.1 AP50.
The results on LVIS~\cite{LVIS} OVD benchmark is shown in \cref{tab:ovd_lvis}.

Note that among the baseline methods, VL-PLM~\cite{vlplm}, ViLD~\cite{vild} and OV-DETR~\cite{OVDETR} use novel class names during training in order to recognize the potential novel objects and assign pseudo labels for them. Consequently, a new detector needs to be trained whenever there is a new set of categories to be detected.
CORA can generalize to any combination of novel categories once trained without tedious re-training.
Other compared methods relying on CLIP require image tag annotations extracted from language descriptions~\cite{medet, bridging, regionclip} or image labels~\cite{detic} during training.
We argue that the extra annotations do not provide additional information over CLIP.
Instead, they serve as a media to transfer the knowledge from CLIP to the detector.
Our method directly adapts the CLIP model to obtain region classifier, thus no extra image-text data is needed.

In this work, both region prompting and anchor pre-matching aim to generalize the knowledge learned from base classes to novel classes.
Consequently, the performance gap between novel and base classes is significantly lowered than the compared methods. 
Note that in OVD, the performance is evaluated by the generalization capability towards novel classes, rather than the bases classes on which they are trained.

\begin{table}[]
\centering
\begin{tabular}{l|lcc}
\toprule
Method       & CLIP model & Novel & Base                        \\
\midrule
CLIP         & RN50       & 58.2  & {\color[HTML]{C0C0C0} 58.6} \\
CLIP-Adapter~\cite{clipadapter} & RN50       & 63.0  & {\color[HTML]{C0C0C0} 80.6} \\
CoOp~\cite{coop}         & RN50       & 64.4  & {\color[HTML]{C0C0C0} 75.7}  \\
CORA         & RN50       & 65.1  & {\color[HTML]{C0C0C0} 70.0} \\
\midrule
CLIP         & RN50x4     & 63.9  & {\color[HTML]{C0C0C0} 62.7} \\
CORA         & RN50x4     & 74.1  & {\color[HTML]{C0C0C0} 76.0} \\
\bottomrule
\end{tabular}
\caption{Results on the region classification task evaluated in mAP. We compare our method with the original CLIP regional classifier and other baseline methods.}
\label{tab:region_prompting}
\end{table}
\subsection{\textcolor{black}{Effectiveness of Region Prompting}}
\label{sec:region_prompt}
\noindent\textbf{Region Prompting}. Since object classification is decoupled from localization in this work, the CLIP-based region classifier can be directly evaluated by the region classification task.
After trained on the base-class annotations of COCO dataset, we evaluate the CLIP-based region classifier on both base and novel classes in the validation set.
We use mean average precision (mAP) as our evaluation metric.

\cref{tab:region_prompting} shows our main result.
Directly evaluating on the CLIP model without further training already achieves considerable performance of 58.2 mAP on the novel classes.
We compare our result with two competitive methods from the adapter and prompt tuning literatures.
CLIP-Adapter~\cite{clipadapter} adopts an additional bottleneck layer to learn new features and performs residual style feature blending with the original pre-trained features. 
CoOp~\cite{coop} prepends shared learnable prompts to the text embeddings before encoded by the CLIP text encoder.
Experiments show that the compared methods bias strongly towards base classes.
Note that the baseline methods adapts the CLIP model by tuning the text input or output feature, which are irrelevant or distant to the regional feature, where the mismatch between regional and whole-image feature occurs.
On the contrary, region prompting directly prompts the mismatched features, thus generalizes better to the novel classes, achieving a performance gain of 6.9 mAP over CLIP on the novel classes.
Region prompting also scales with larger backbones. 
On the RN50x4 CLIP backbone, region prompting further boosts CLIP by 10.2\% mAP over the corresponding CLIP model.

\begin{table}[]
\centering
\begin{tabular}{l|lcc}
\toprule
CLIP model       & Feature & Novel & Base                        \\
\midrule
RN50         & whole-image       & 43.8  & 40.5 \\
RN50 & regional       & 58.6  & 58.2 \\
\bottomrule
\end{tabular}
\caption{Comparison of the whole-image classification pipeline and the regional classification pipeline. Results are reported in mAP.}
\label{tab:whole_image}
\end{table}
\noindent\textbf{Comparison with whole-image classification. }
A common practice of region classification by CLIP is to crop the regions and classify them as separate images.
We compare the region classifier with the common practice on the original CLIP weights without region prompting.
As shown in \cref{tab:whole_image}, when classifying a region as a whole-image, the performance is significantly lower than using the regional feature, despite the extra computation.
We attribute the performance gap to the missing context of cropped images.

\subsection{\textcolor{black}{Effectiveness of Anchor Pre-Matching}}
\label{sec:anchor_pre-matching}
We conduct ablation studies to validate anchor pre-matching and the proposed training techniques.

\begin{table}[]
\centering
\resizebox{.47\textwidth}{!}{
\begin{tabular}{ccc|c}
\toprule
\multicolumn{1}{l}{\multirow{2}{*}{\begin{tabular}[c]{@{}l@{}}Query \\ Conditioning\end{tabular}}} & \multicolumn{1}{l}{\multirow{2}{*}{\begin{tabular}[c]{@{}l@{}}Per-Class\\ Post-Matching\end{tabular}}} & \multicolumn{1}{l|}{\multirow{2}{*}{\begin{tabular}[c]{@{}l@{}}CLIP-Aligned\\ Labeling\end{tabular}}} & \multirow{2}{*}{Novel} \\
\multicolumn{1}{l}{}                                                                               & \multicolumn{1}{l}{}                                                                                   & \multicolumn{1}{l|}{}                                                                                 &                        \\
\midrule
\xmark                                                                                             & \xmark                                                                                                 & \cmark                                                                                                & 26.0                   \\
\cmark                                                                                             & \xmark                                                                                                 & \cmark                                                                                                & 29.9                   \\
\cmark                                                                                             & \cmark                                                                                                 & \xmark                                                                                                & 40.9                  \\
\cmark                                                                                             & \cmark                                                                                                 & \cmark                                                                                                & \textbf{41.7}          \\
\bottomrule
\end{tabular}
}
\caption{Ablation studies on anchor pre-matching and CLIP-aligned labeling. Anchor pre-matching consists of query conditioning and per-class post-matching. }
\label{tab:abl_apm}
\end{table}
\noindent\textbf{Anchor Pre-Matching. }
Anchor pre-matching consists of two separate operations: query conditioning and per-class post-matching.
We analyze their effects in \cref{tab:abl_apm}.
Firstly, we train a model without anchor pre-matching. We keep using DAB-DETR\cite{dabdetr} as the localizer, and use the same number of object queries and anchor boxes for fair comparison.
The object queries are not pre-matched to classes, and are initialized as 0 as done in DAB-DETR.
Since the model predictions are not pre-matched, the per-class post-matching is replaced by the vanilla one-to-one matching mechanism in DETR.
Then, we classify the anchor boxes and condition the object queries on the class name embeddings, but do not set constraint on the post-matching, which boosts the performance on novel classes by 3.9 AP50.
After adopting the full anchor pre-matching, the performance is significantly boosted by 10.8 AP50.

\begin{figure}
  \centering
    \hspace{0.cm}
    \includegraphics[width=1.\linewidth]{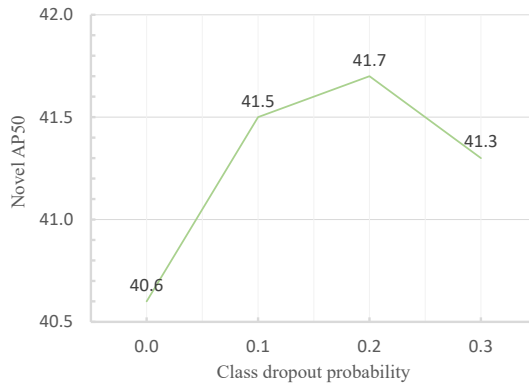}
    \vspace{-1.2cm}
  \caption{Different choices of $p$ for class dropout. }
  \label{fig:class_dropout}
\end{figure}

\noindent\textbf{Training Techniques. }
Class dropout and CLIP-aligned labeling are two training techniques that help the model generalize better.
In \cref{fig:class_dropout}, we examine the effect of different dropout probability.
We find that the models trained with class dropout consistently outperform the baseline, and $p=0.2$ gives the best performance.
In \cref{tab:abl_apm}, we validate the effectiveness of CLIP-Aligned Labeling.

\section{Conclusion}
The core challenge in open-vocabulary detection is how to effectively transfer the knowledge learned from the base classes to the unseen novel classes for evaluation.
In this work, we directly adapt the  CLIP into a region classifier, and mitigate the distribution gap between whole-image feature and regional feature through region prompting, which successfully generalizes to the novel categories.
Different from prior works that rely on a fixed RPN for novel class localization, we achieve efficient class-aware localization by the proposed anchor pre-matching mechanism.
Experiments show that our method can better transfer the knowledge from base classes to unseen novel classes with a smaller gap than prior works.
We hope our work can help other researchers gain better insight on the OVD problem and develop better open-vocabulary detectors.

\noindent\textbf{Acknowledgement.}
This project is funded in part by National Key R\&D Program of China Project 2022ZD0161100, by the Centre for Perceptual and Interactive Intelligence (CPII) Ltd under the Innovation and Technology Commission (ITC)'s InnoHK, by General Research Fund of Hong Kong RGC Project 14204021. 
Hongsheng Li is a PI of CPII under the InnoHK.
This project is also supported by SenseTime Collaborative Research Grant.

{\small
\bibliographystyle{ieee_fullname}
\bibliography{egbib}
}
\clearpage
\appendix
\section{Extra Implementation Details}
We set the exponential moving average factor as 0.99996. 
We follow prior works in the DETR literature to apply auxiliary loss on the output of each intermediate decoder layer, but anchor pre-matching is conducted only once.
The hyper-parameters of the matching cost are identical to the corresponding loss coefficients.
During training, we use random flip, random resize and random crop to augment the input image.
The smaller edge of an input image is resized to a value between 480 and 800, and the larger edge is resized while keeping the aspect ratio.
The maximum length of the resized larger edge is 1333.
During inference, the temperature $\tau$ of the classification logits is set to 0.01.
For image resolution during inference, the smaller edge is resized to 800, and the larger edge is resized accordingly by keeping the aspect ratio. 
The image is further resized when necessary to make sure the larger edge is no longer than 1333.
During inference, we multiply the logit of novel classes by a factor of 8.0.

\section{Localization Capability of CORA}
In the main text, we demonstrate the effectiveness of our method by evaluating it on both region classification and object detection.
In this section, we further show the superior novel object localization capability of our method.

In order to evaluate the localization capability, we only take the predicted box coordinates from the model for evaluation.
The class label of each box is assigned by the ground truth box with highest IoU.
The confidence score is replaced by the highest IoU.
These modifications eliminate the effect of the classifier, and make sure that only the localization capability is evaluated.
The predictions are then evaluated on the standard COCO OVD benchmark.

\begin{table}[h]
\centering
\begin{tabular}{l|cc}
\toprule
Method        & Novel & {\color[HTML]{C0C0C0}Base}                        \\
\midrule
RegionCLIP               &  81.7 & {\color[HTML]{C0C0C0}88.2} \\
CORA     & 83.4  & {\color[HTML]{C0C0C0}88.1} \\
\bottomrule
\end{tabular}
\caption{Comparison on the localization capability. The performance is evaluated in AP50.}
\label{tab:loc}
\end{table}
The result is shown in ~\cref{tab:loc}.
We compare our method with RegionCLIP, which is a strong baseline on the COCO OVD benchmark.
The novel-to-base performance gap of CORA is significantly lower than the baseline, demonstrating a better generalization capability towards the novel classes.

\begin{figure}
  \centering
  \begin{subfigure}{0.95\linewidth}
    \includegraphics[width=1.\linewidth]{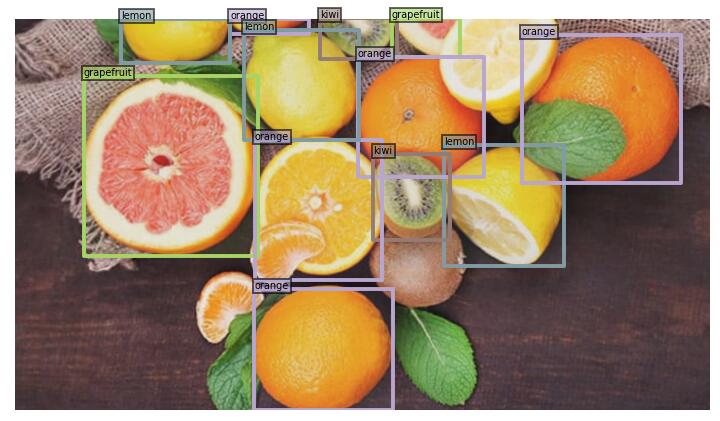}
    \caption{Categories: kiwi, banana, orange, grapefruit, lemon.}
    \label{fig:fruit}
  \end{subfigure}
  \hfill
  \begin{subfigure}{0.95\linewidth}
    \includegraphics[width=1.\linewidth]{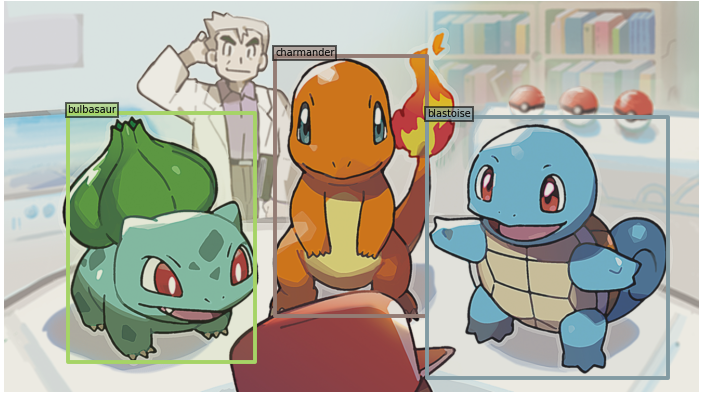}
    \caption{Categories: Bulbasaur, Charmander, Blastoise, Torchic, Treecko.}
    \label{fig:pokemon}
  \end{subfigure}
  \hfill
  \begin{subfigure}{0.95\linewidth}
    \includegraphics[width=1.\linewidth]{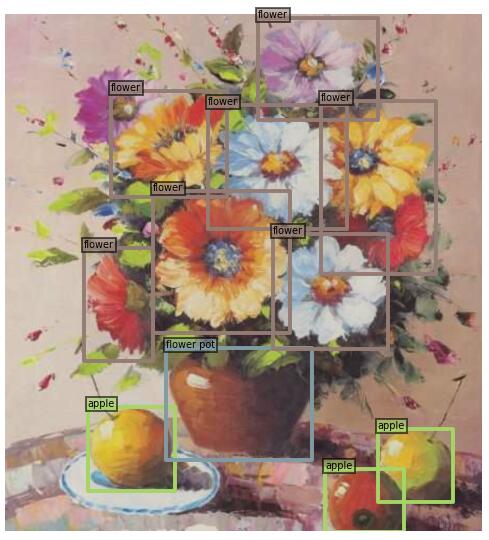}
    \caption{Categories: flower, apple, flower pot.}
    \label{fig:oil}
  \end{subfigure}
  \caption{Visualization of predictions on base and novel classes.}
  \label{fig:vis}
\end{figure}

\section{Visualizations}
\cref{fig:vis} visualizes the predictions of CORA on images with novel objects. 


\end{document}